\documentclass[letterpaper, 10 pt, conference]{ieeeconf}  
\IEEEoverridecommandlockouts                    

\RequirePackage{fixltx2e}

\newcommand{\vect}[1]{\mathbf{#1}}

\newcommand{\diffs}[3]{\frac{\partial^2 #1}{
\ifx#2#3 
\partial #2^2
\else
\partial #2 \partial #3
\fi
}}

\usepackage[usenames,dvipsnames]{xcolor}

\let\labelindent\relax
\usepackage[inline]{enumitem}
\setlist*[enumerate,1]{label=\it{\arabic*)}}

\usepackage{tikz}
\usetikzlibrary{calc}
\usetikzlibrary{matrix}
\usetikzlibrary{chains}
\usetikzlibrary{shapes.geometric}
\usetikzlibrary{arrows.meta}
\usetikzlibrary{decorations.markings}
\usetikzlibrary[arrows.meta,bending]

\usepackage{comment}

\usepackage{blkarray}

\usetikzlibrary{calc}

\makeatletter

\let\proof\@undefined
\let\endproof\@undefined
\makeatother

\usepackage{amsmath} 
\usepackage{amssymb}  
\usepackage{amsthm}
\usepackage{amsfonts}
\usepackage{mathtools}

\usepackage{verbatim}
\usepackage[outdir=./epsToPdf]{epstopdf}
\usepackage{graphicx}
\usepackage{caption}
\usepackage{subcaption}
\usepackage{epstopdf}

\usepackage{soul} 

\usepackage{times}

\usepackage[framemethod=TikZ]{mdframed}

\usepackage{siunitx}

\usepackage[ruled,vlined,linesnumbered]{algorithm2e}

\usepackage[us]{datetime}
\usepackage{caption}
\usepackage{hyperref} 
\hypersetup{
colorlinks,
citecolor=black,
filecolor=black,
linkcolor=black,
urlcolor=black,
pdfauthor={},
pdfsubject={},
pdftitle={}
}

\usepackage{todonotes}

\usepackage{multicol}

\usepackage{latexsym}

\usepackage{cite}

\usepackage{caption}

\usepackage[normalem]{ulem}

\definecolor{myblue}{rgb}{.8, .8, 1}

\usepackage{amsmath}
\usepackage{empheq}

\newlength\mytemplen
\newsavebox\mytempbox

\makeatletter
\newcommand\mybluebox{%
    \@ifnextchar[
       {\@mybluebox}%
       {\@mybluebox[0pt]}}

\def\@mybluebox[#1]{%
    \@ifnextchar[
       {\@@mybluebox[#1]}%
       {\@@mybluebox[#1][0pt]}}

\def\@@mybluebox[#1][#2]#3{
    \sbox\mytempbox{#3}%
    \mytemplen\ht\mytempbox
    \advance\mytemplen #1\relax
    \ht\mytempbox\mytemplen
    \mytemplen\dp\mytempbox
    \advance\mytemplen #2\relax
    \dp\mytempbox\mytemplen
    \colorbox{myblue}{\hspace{1em}\usebox{\mytempbox}\hspace{1em}}}

\usepackage{verbatim}
\mdfdefinestyle{MyFrame}{%
  linecolor=blue,
  outerlinewidth=2pt,
  roundcorner=10pt,
  innertopmargin=2pt,
  innerbottommargin=2pt,
  innerrightmargin=10pt,
  innerleftmargin=10pt,
  backgroundcolor=gray!50!white
}

\surroundwithmdframed[style=MyFrame]{macb}

\title{\LARGE \bf Hand-worn Haptic Interface for Drone Teleoperation}

\author{
Matteo Macchini, Thomas Havy, Antoine Weber, Fabrizio Schiano, and Dario Floreano
\thanks{The authors are with the Laboratory of Intelligent Systems, École Polytechnique Fédérale de Lausanne, CH-1015 Lausanne (EPFL), Switzerland.}%
}

\usepackage{cleveref}

\newcommand{\kron}{\otimes}

\renewcommand{\vect}[1]{\boldsymbol{#1}}
\newcommand{\matr}[1]{\boldsymbol{#1}}

\newcommand{\zeros}[2]{
\ifthenelse{\equal{#2}{1}}{\vect{0}_{#1}}{\matr{\cancel{O}}_{#1 \times #2}}
}
\newcommand{\ones}[2]{
\ifthenelse{\equal{#2}{1}}{\vect{1}_{#1}}{\matr{1}_{#1 \kron #2}}
}

\newcounter{simulationcase}

\crefname{simulationcase}{case}{cases}
\Crefname{simulationcase}{Case}{Cases}

\crefname{equation}{}{}
\Crefname{equation}{Equation}{Equations}
\crefmultiformat{equation}{(#2#1#3}%
  {--#2#1#3)}{, (#2#1#3)}{ and~(#2#1#3)}
\Crefmultiformat{equation}{Equations (#2#1#3}%
  {--#2#1#3)}{, (#2#1#3)}{ and~(#2#1#3)}
\crefname{figure}{Fig.}{Figs.}
\Crefname{figure}{Figure}{Figures}
\crefname{part}{Part}{Parts}
\Crefname{part}{Part}{Parts}
\crefname{chapter}{Chapt.}{Chapts.}
\Crefname{chapter}{Chapter}{Chapters}
\crefname{section}{Sect.}{Sects.}
\Crefname{section}{Section}{Sections}
\crefname{subsection}{Sect.}{Sects.}
\Crefname{subsection}{Section}{Sections}
\crefname{subsubsection}{Sect.}{Sects.}
\Crefname{subsubsection}{Section}{Sections}
\crefname{appsec}{Appendix}{Appendices}
\Crefname{appsec}{Appendix}{Appendices}
\crefname{appendix}{Appendix}{Appendices}
\Crefname{appendix}{Appendix}{Appendices}
\crefname{subappendix}{Appendix}{Appendices}
\Crefname{subappendix}{Appendix}{Appendices}
\crefname{lemma}{Lemma}{Lemmas}
\Crefname{lemma}{Lemma}{Lemmas}
\crefname{remark}{Remark}{Remarks}
\Crefname{remark}{Remark}{Remarks}
\crefname{prob}{Problem}{Problems}
\Crefname{prob}{Problem}{Problems}
\crefname{constr}{Constraint}{Constraints} 
\Crefname{constr}{Constraint}{Constraints}
\crefname{algorithm}{Algorithm}{Algorithms}
\Crefname{algorithm}{Algorithm}{Algorithms}
\crefname{prop}{Prop.}{Props.}
\Crefname{prop}{Proposition}{Propositions}
\crefname{ALC@unique}{step}{steps}
\Crefname{ALC@unique}{Step}{Steps}

\renewcommand\vect[1]{{\boldsymbol{#1}}}

\renewcommand{\mod}[1]{\textcolor{NavyBlue}{#1}}
\newcommand{\removed}[1]{\textcolor{NavyBlue}{\sout{#1}}}
\newcommand{\rem}[1]{\textcolor{NavyBlue}{\sout{#1}} \textcolor{red}{ remove?}}
\renewcommand{\removed}[1]{}
\renewcommand{\rem}[1]{}
\renewcommand{\mod}[1]{#1}
\definecolor{mygray}{gray}{0.75} 
\newcommand{\old}[1]{}  

\allowdisplaybreaks

\begin{document}
\maketitle

\begin{abstract}
Drone teleoperation is usually accomplished using remote radio controllers, devices that can be hard to master for inexperienced users.
Moreover, the limited amount of information fed back to the user about the robot's state, often limited to vision, can represent a bottleneck for operation in several conditions.
In this work, we present a wearable interface for drone teleoperation and its evaluation through a user study. 
The two main features of the proposed system are a data glove to allow the user to control the drone trajectory by hand motion and a haptic system used to augment their awareness of the environment surrounding the robot.
This interface can be employed for the operation of robotic systems in line of sight (LoS) by inexperienced operators and allows them to safely perform tasks common in inspection and search-and-rescue missions such as approaching walls and crossing narrow passages with limited visibility conditions.
In addition to the design and implementation of the wearable interface, we performed a systematic study to assess the effectiveness of the system through three user studies (n = 36) to evaluate the users' learning path and their ability to perform tasks with limited visibility.
We validated our ideas in both a simulated and a real-world environment.
Our results demonstrate that the proposed system can improve teleoperation performance in different cases compared to standard remote controllers, making it a viable alternative to standard Human-Robot Interfaces.

\end{abstract}

\vspace{4mm}

Supplementary video: \href{https://youtu.be/ol7UT1ApLpM}{https://youtu.be/ol7UT1ApLpM}

\section{Introduction}
Despite the remarkable advancements in the autonomy level of industrial and service robotic systems, the ability for robots to be fully autonomous is still a far goal~\cite{gibo_shared_2016}.
Telerobotics, the branch of robotics studying systems in which a human partially or fully operates a robot, still finds predominant applications for tasks such as navigation in challenging environments, or minimally invasive surgery~\cite{diftler_robonaut_2011, khatib_ocean_2016,bodner_first_2004}. 
As a consequence, the need for intuitive and efficient Human-Robot Interfaces (HRIs) is vital to allow a larger population of individuals to control these robots, a task typically restricted to highly trained professionals \cite{casper_human-robot_2003, chen_supervisory_2011}.
Body-Machine Interfaces (BoMIs), a subfield of HRIs, raised relevant interest in the field of robotics in recent years as they find foundation in the intuitive control humans can exert on their body motion \cite{casadio_body-machine_2012}.
Moreover, wearable technologies allow for a closer physical bond between the human and the machine, which can be strengthened through the sense of touch. 
Haptic interfaces are notably capable of increasing the user's awareness of the robot's state and improve teleoperation performance \cite{hannaford_performance_1991, siciliano_haptics_2008, aggravi2018design}.
The combination of these two components, body motion for control and haptics for state feedback, is a fascinating, still partially unexplored technique that is showing promising perspective for telerobotics in a multitude of fields \cite{rognon_flyjacket:_2018, rognon_soft_2019, diftler_robonaut_2011, wang_hermes_2015,bimbo2017teleoperation}.

Small drones are attracting considerable and growing interest in research labs with perspectives in several applications \cite{sesar_european_2016}.
These applications include, but are not limited to, transportation, communication, agriculture, disaster mitigation, and environment preservation \cite{floreano_science_2015}. 
For drone teleoperation, commonly employed interfaces can represent a non-intuitive and sophisticated control device for naive users. This kind of interfaces requires a high concentration and cognitive effort for long-term use \cite{peschel_humanmachine_2013}.

\begin{figure}[t]
\includegraphics[width=\columnwidth]{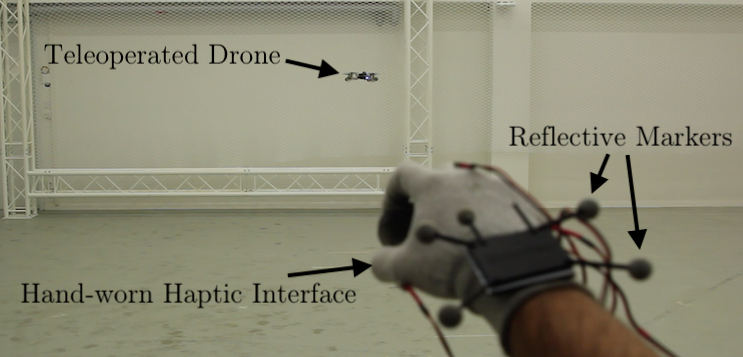}
	\caption{System Setup. A user controls the motion of a quadcopter through the proposed wearable haptic interface.}
	\label{f:intro}
\end{figure}

Motion-based systems have already been successfully implemented and demonstrated to be superior to remote controllers for aerial robots \cite{miehlbradt_data-driven_2018, macchini_personalized_2020}.
Haptic feedback has also been employed both to increase operational performance and represent air pressure during flight for enhanced immersion \cite{rognon_flyjacket:_2018, rognon_soft_2019}. 
Several solutions have been proposed, mainly based on a set of discrete gestures to control the flight of a quadrotor in the user's line of sight (LOS).
Motion-based systems for flying robots have been implemented using stereo vision devices for tracking the motion of the operator's full body or their hands \cite{ikeuchi_kinecdrone:_2014, sanna_kinect-based_2013,sarkar_gesture_2016}.
The restriction to a predefined vocabulary of command inputs, such as “take off”, “go left”, despite alleviating the user from continuous interaction with the robot, is not suited for fine control of its trajectory.
A BoMI based on pointing gestures for drone teleoperation has shown comparable performance with respect to joysticks for landing maneuvers \cite{gromov_proximity_nodate}.
However, this work is limited to a motion on a plane, and cannot thus cope with 3D trajectories.
Hand-worn interfaces have also been considered for the control of quadrotor teams, both based on muscular activity and, recently in a pioneering work on motion \cite{stoica_remote_2014,tsykunov_swarmtouch:_2018,tsykunov2019swarmtouch, ibrahimov2019dronepick}.
Additionally, Tsykunov et al., in \cite{tsykunov2019swarmtouch,tsykunov_swarmtouch:_2018}, used vibrotactile feedback on the user's hand to represent the drone swarm behaviour.
This approach, though, includes autonomous obstacle avoidance on a 2D surface, through a motion capture system, making the user passive towards this task.
The use of vibrating tactors has also been proposed for drone obstacle avoidance in 2D, in a simulated environment \cite{hasegawa_comparison_2016}.

Despite the recent scientific and industrial interest for motion-based interfaces and haptic feedback for the control of aerial robots, a study of the real usability of such systems in real-world cases is still lacking.
To fill this gap, we present a systematic approach to provide more insights on how effective is controlling a drone through hand motion and the role of haptics in augmenting the performance in particular tasks, which are fundamental in several scenarios such as search-and-rescue or inspection \cite{jimenez2019contact,falanga2018foldable,delmerico2019current}.
In this paper, we first investigate the effects of a paradigm shift for robot teleoperation consisting of mapping the position of the user's hand into a position command for the robot, as in \cref{f:intro}.
Towards this goal, we developed a wearable interface based on a haptic data glove.
We want not only to inform the user about the state of the drone but augment their exteroceptive capabilities with haptic cues to allow a finer and safer operation of the robot.
Generally, none of the above solutions can collect environmental information from onboard sensors. 
Instead, we equipped the drone with six distance sensors to allow onboard obstacle detection.
This makes our system a viable solution for 'non-trivial' environments
(e.g., inside buildings, in urban canyons) where centralized sensing facilities, such as GNSS, are not available, and the robots can only rely on local sensors.

Finally, through user studies, we evaluated the human learning capabilities for the developed wearable HRI first without the haptic feedback, and,  subsequently, with the haptic feedback to test its effectiveness in a navigation task in cluttered environments. To our knowledge, our solution proposes the first haptic device capable of representing the presence of obstacles in the three dimensions on the same wearable system used for teleoperation, and at the same time, the first approach based on onboard sensing.
We validated our ideas with both simulation and experiments with real quadrotors. 

\section{Method}
In this section, we present the proposed human-robot interface, focusing on three main parts: the motion-based controller allowing the user to control a drone in simulation by hand motion, the wearable haptic device used to represent the presence of obstacles through vibrotactile feedback, and the implementation on a real quadrotor.

\textbf{Motion-based controller}: 
the first part of our implementation consists of a wearable interface used to control a quadrotor remotely.
The 3D position of the operator's hand is tracked with a motion capture system (MoCap)\footnote{\href{https://optitrack.com/}{https://optitrack.com/}}.
The MoCap streams the hand position at a rate of $120Hz$ to a drone simulator implemented in Unity3D.
The simulator is based on a third-party model\footnote{Berkeley drone simulator GitHub:  \href{https://github.com/UAVs-at-Berkeley/UnityDroneSim}{https://github.com/UAVs-at-Berkeley/UnityDroneSim}} which reproduces the dynamics of a quadrotor and the corresponding attitude control system.
A PID controller was used to control the position of the robot.
The simulation environment was designed to reproduce the MoCap room appearance and size.
The goal position is set to correspond to the hand position, with a scaling factor of $8$, chosen to match the user's arm reachable space with the size of the room. 
\mod{We included a clutch mechanism to activate and deactivate the interface. 
The interface is activated by pressing the left button of a mouse, and the drone will move according to the user's hand position. 
Releasing the button will allow the user to move freely without affecting the drone motion.}
At the start of the simulation, the goal position is reset to the drone's initial location.

\begin{figure}[h]
\begin{subfigure}{.5\columnwidth}
\centering
\includegraphics[height=5cm]{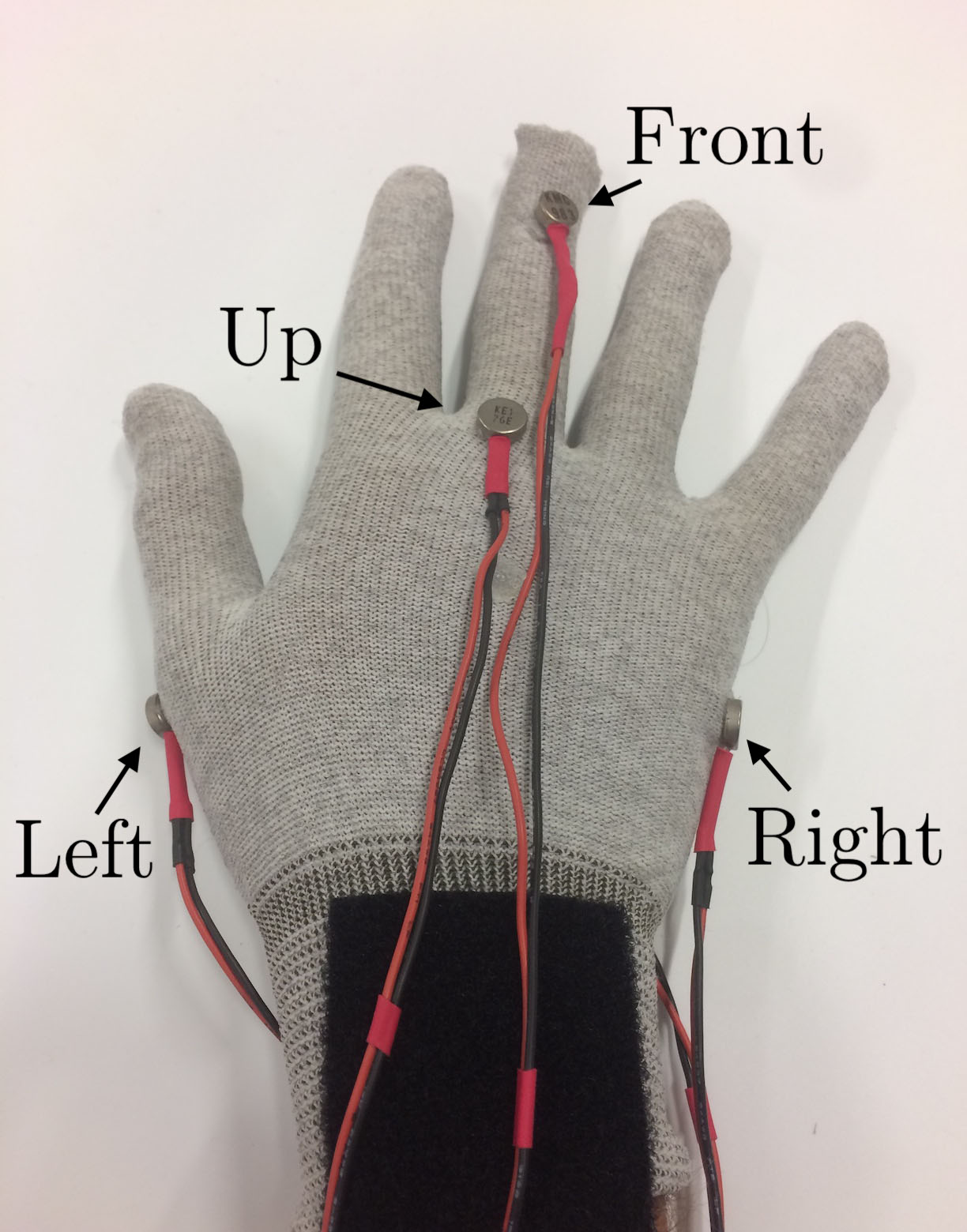}
\end{subfigure}%
\begin{subfigure}{.5\columnwidth}
\centering
\includegraphics[height=5cm]{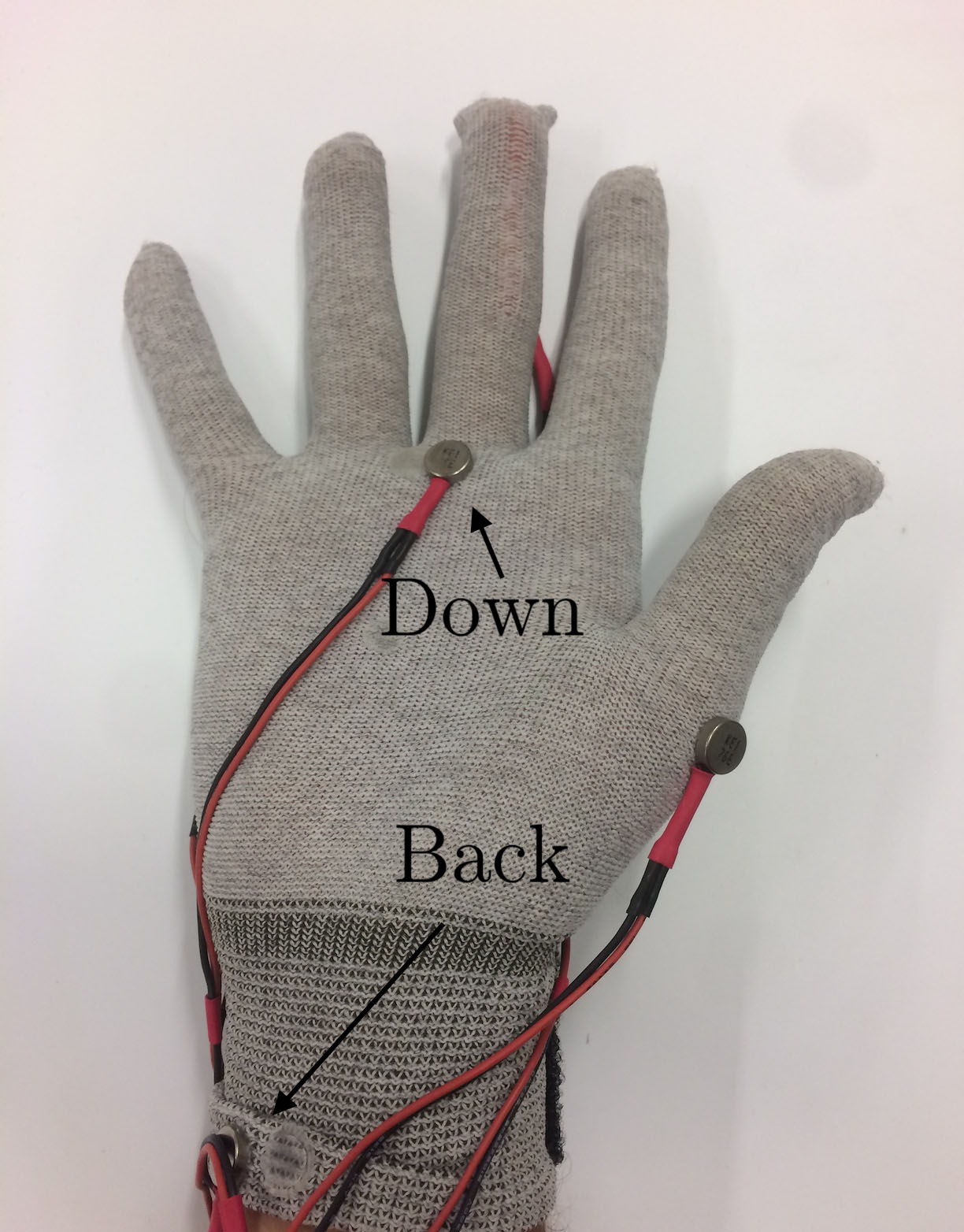}
\end{subfigure}
\caption{Location of tactors on the upper (left) and lower (right) side of the wearable teleoperation interface.}
\label{f:glove}
\end{figure}

\begin{figure}[h]
\begin{subfigure}{.5\columnwidth}
\centering
\includegraphics[height=3.3cm]{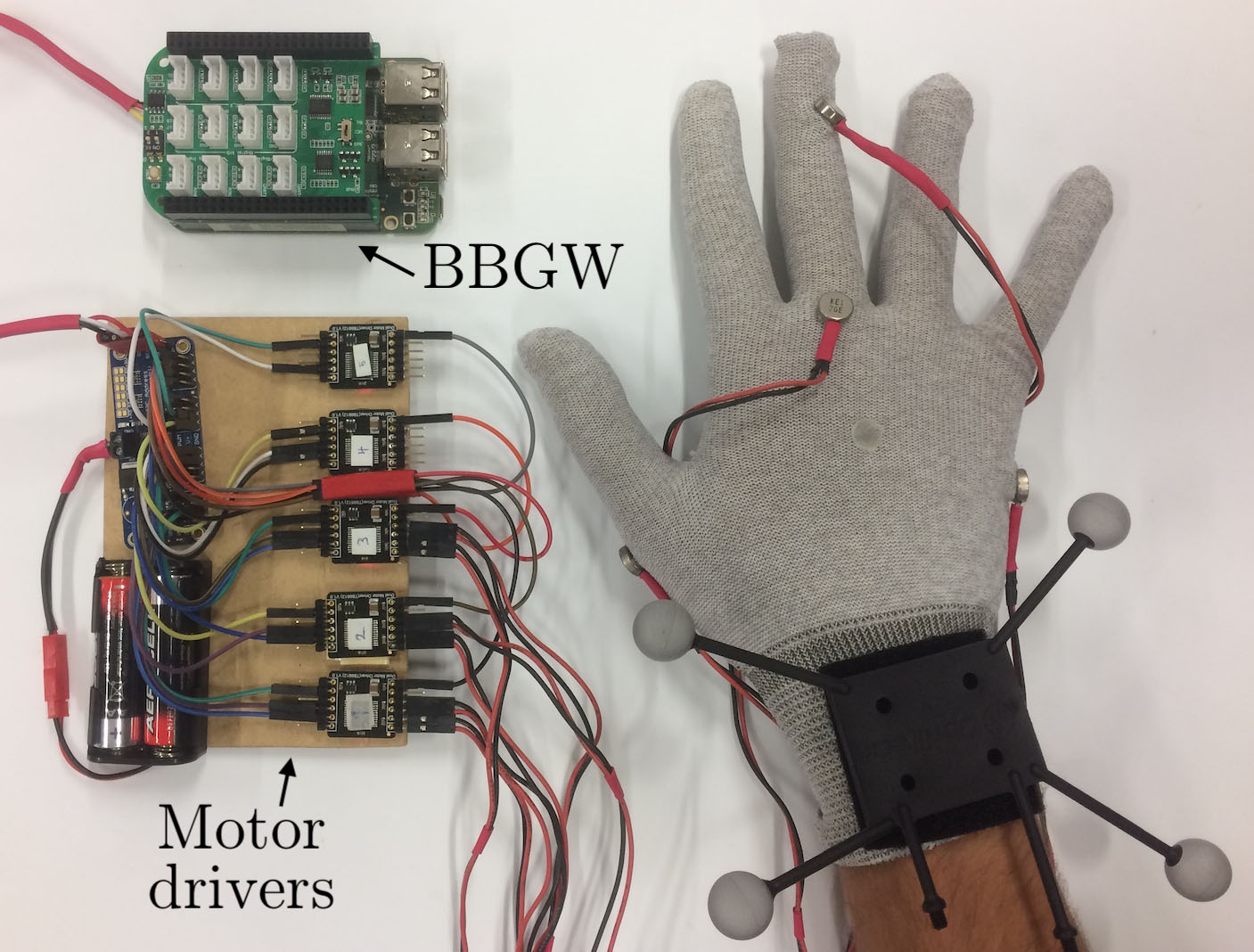}
\end{subfigure}%
\begin{subfigure}{.5\columnwidth}
\centering
\includegraphics[height=3.3cm]{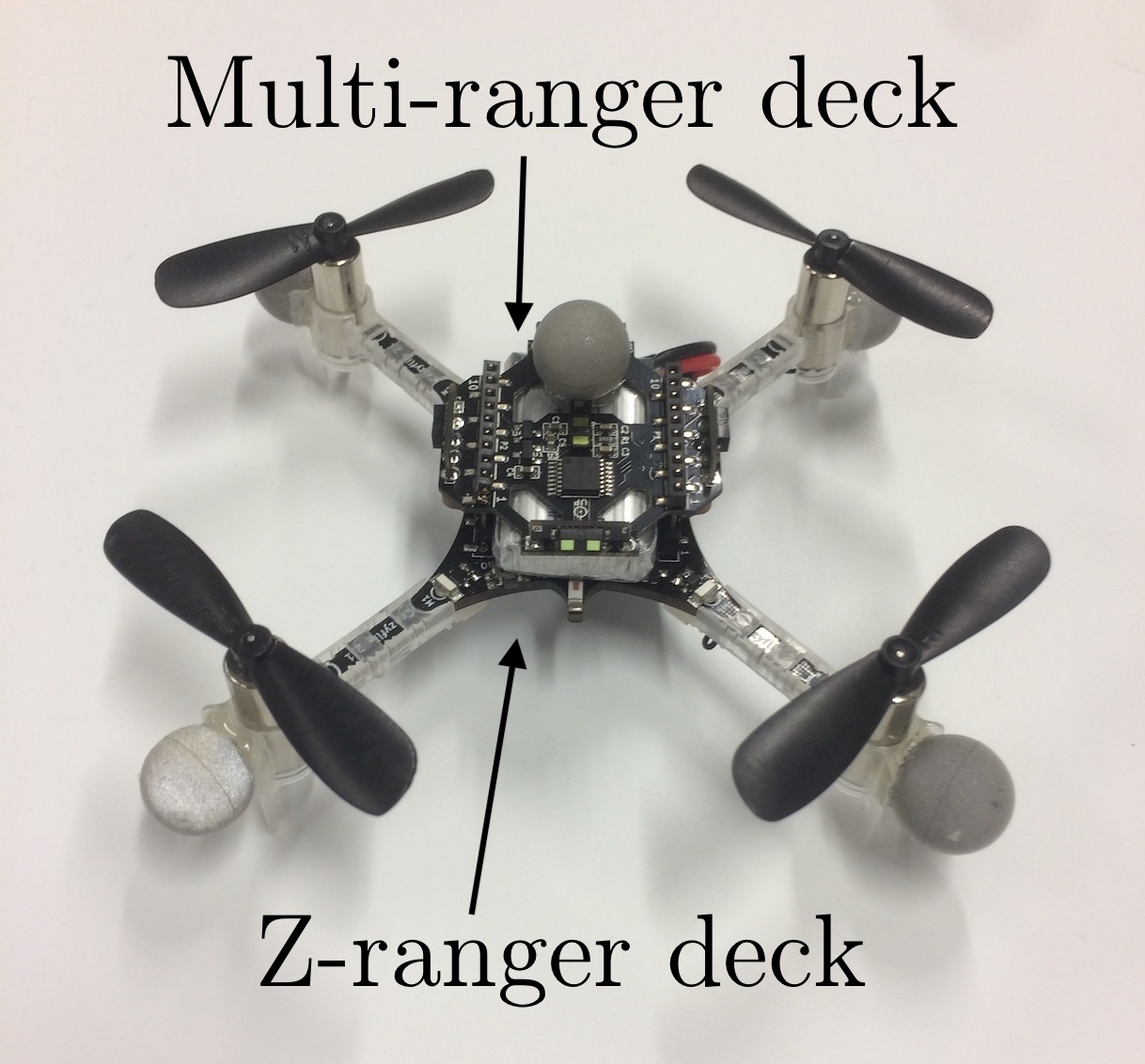}
\end{subfigure}
\caption{Wearable interface with hardware control device (left) and CrazyFlie 2.1 equipped with proximity sensors (right).}
\label{f:system}
\end{figure}

\textbf{Wearable haptic device}:
the second component of the presented solution is a haptic glove, capable of streaming to the user information about the environment surrounding the robot.
The glove embeds six tactors, corresponding to each axis of motion of the quadrotor (up-down, left-right, front-back), placed as in \cref{f:glove}. 
This configuration allows for omnidirectional coverage.
Each tactor is implemented using a vibrating motor, installed on the external surface of a fabric glove.
We programmed a BeagleBone Green Wireless board to interface with an H-bridge driver to regulate the vibration of the six tactors as in \cref{f:system} (left).
When the quadrotor is located near an obstacle, the tactor installed on the corresponding direction will vibrate with an intensity proportional to the obstacle's proximity.
We can express the vibration intensity $i$ as $i = MT/d$ where $M$ is the maximum possible intensity, $T$ an arbitrary threshold, set to $0.5m$, and $d$ the obstacle distance. To estimate the distances, we implemented in the simulator a model of an omnidirectional laser sensor, installed on the drone, reproducing the characteristics of a commercial device\footnote{Crazyflie multiranger deck: \href{https://www.bitcraze.io/multi-ranger-deck/}{https://www.bitcraze.io/multi-ranger-deck/}}.

\textbf{Hardware Implementation}:
as a final step, we implemented the wearable interface on a real quadrotor to test its real-world capabilities.
The platform of choice was the centimeter-scale Crazyflie 2.1 drone \cite{Crazyflie}.
This quadrotor is well suited for testing the newly designed interface, for its robustness and simplicity of deployment. 
Moreover, being small-sized and lightweight, these platforms are\removed{well-suited} \mod{indicated} for indoor flight in limited space, such as a Motion Tracking Hall.
The Crazyflie platform supports a set of expansion decks, designed for different use.
For this paper, we equipped the drone with two different decks, the multi-ranger deck, and the Z-ranger deck, providing proximity measurements on the three axes. For all these reasons, and because it comes with open source software and hardware, this platform recently gained much attention among researchers \cite{honig2018trajectory,gabrich2018flying}. It is shown in  \cref{f:system} (right).
The selected sensors are the same that we modeled in the simulator, as described in the previous section.
The quadrotor receives control commands from a ground station on a dedicated radio channel through the manufacturer's official radio interface.
The radio communication provides position setpoints to the drone and receives its state and sensor information.
The drone position is tracked through an Optitrack motion capture system and controlled through a PID controller.
We used the Crazyswarm library to implement the control system through ROS  \cite{preiss_crazyswarm:_2017}.

\section{Results}
In this section, we summarize the experimental protocols and results obtained through our study\footnote{The experiments were approved by the \'Ecole Polytechnique F\'ed\'erale de Lausanne Human Research Ethics Committee.}.
We conducted both extensive user tests in simulation and qualitative ones in hardware.
We always show results for the simulation phase in the top row and results from the hardware implementation in the bottom of the figures.
All the performed experiments consist of a statistical analysis of the impact of using the proposed interface on human subjects \removed{(hereafter ``Group W'')}compared with the use of a standard remote controller \removed{(hereafter ``Group R'')}.
\mod{We consider 3 experimental conditions, corresponding to the use of 3 different interfaces:
\begin{itemize}
    \item A standard remote controller (hereafter ``R'')
    \item The wearable interface without haptic feedback \\(hereafter ``W'')
    \item The wearable interface with haptic feedback \\(hereafter ``H'')
\end{itemize}}
We employed the Kruskal-Wallis test to assess the significance of our results (T-test) and the Levene test to assess the variances equality (F-test) \cite{kruskal_use_1952,brown_robust_1974}.
Statistical significance was tested only on the simulation results due to the limited number of participants who took part in the hardware experiment.
\mod{
First, we tested the users' perception of the haptics component of the interface. 
In a second step, we evaluated the subject's learning capabilities of the motion-based interface.}
Later, we proposed a practical case to appreciate the effectiveness of the haptic interface.
It is important to remark that many participants had prior experience using remote controllers \mod{($7.28/10 \pm 2.24$)}, while all of them tested our wearable system for the first time.

\textbf{I - Validation of the haptic glove: }
The validation for the design of the haptic glove consisted of two different experiments.
The first experiment was performed to determine the mean response time of different subjects to a vibration stimulus. 
5 subjects took part in the experiment.
As they were wearing the tactile glove, a vibrating pulse was randomly sent to one of the 6 different motors. 
The subjects were asked to press the spacebar key on a computer whenever they felt the pulse. 
The response time was recorded, and this entire process was repeated  20 times per subject, recording their response time. 
After each pulse, the system randomly waited between $6$ and $12$ seconds to send the next one.

In a second step, we validated the tactors placement.
First, a sample of all the 6 directional pulses was applied sequentially for 1 second to introduce the subjects to all the different options. 
During this part, the subjects were informed about the pulse direction.
After, a pulse was randomly sent to one of the six motors for a given amount of time, corresponding to the observed average response time. 
After each impulse, the subjects were asked to say which direction they thought the pulse was coming from.
5 subjects experimented 20 pseudo-random pulses.

\textbf{\textit{Haptic cues are easy to decode for operators: }}
In the first phase of this experiment, we identified an average response time of $0.45 \pm 0.086 s$.
The accuracy of the subjects' perception is summarized in the confusion matrix in \cref{t:confusion}.
Overall, $98\%$ of the tactile cues were correctly decoded.

\begin{table}[th]
\begin{center}
\begin{tabular}{| l | c | c | c | c | c | c |}
 \hline
    & \textbf{back} & \textbf{front} &\textbf{left} &\textbf{right} &\textbf{up} &\textbf{down} \\
 \hline
 \textbf{back} & $\textbf{100}$ & $0$ & $0$ & $0$ & $0$ & $0$ \\
 \hline
 \textbf{front} & $0$ & $\textbf{100}$ & $0$ & $0$ & $0$ & $0$ \\
 \hline
 \textbf{left} & $0$ & $0$ & $\textbf{100}$ & $0$ & $0$ & $0$ \\
 \hline
 \textbf{right} & $0$ & $0$ & $0$ & $\textbf{100}$ & $0$ & $0$ \\
 \hline
 \textbf{up} & $0$ & $4.76$ & $0$ & $4.76$ & $\textbf{90.48}$ & $0$ \\
 \hline
 \textbf{down} & $0$ & $0$ & $0$ & $0$ & $0$ & $\textbf{100}$ \\
 \hline
\end{tabular}
\caption{Confusion matrix for the recognition of the haptic cues. Participants were able to distinguish different directions with an overall success rate of $98\%$.}
\label{t:confusion}
\end{center}
\end{table}\

\begin{figure}[h]
\includegraphics[width=\columnwidth]{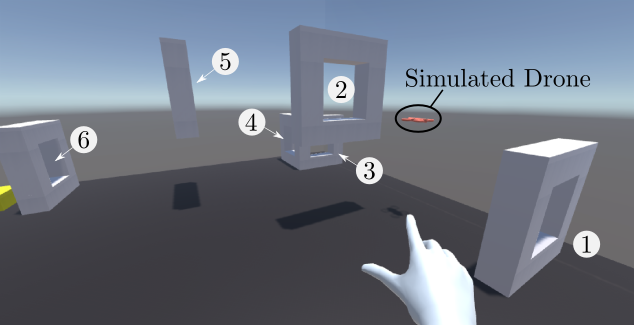}
	\caption{Simulation environment for the learning task. In the figure the user's hand position, the drone and the path composed of 6 obstacles are shown.}
	\label{f:path}
\end{figure}

\textbf{II - Learning the motion-based interface (simulation): }
9 participants were recruited to perform the learning experiment. \mod{We tested condition R vs W.}
The experiment consisted of the repetition of a navigation task where the participants controlled a simulated drone through a path composed of six gates, shown in \cref{f:path}.\removed{Two different conditions were considered: a remote controller and the proposed wearable device.}
The subjects were asked to navigate the quadrotor in simulation through the path five times for two runs, for a total of ten repetitions.
Four subjects started with the remote and the remaining ones with the wearable device in a pseudo-random fashion, in order to balance out possible biasing effects.
After five runs, we let them use the second interface.
All subjects wore a Head-Mounted Display (HMD) and performed the task in a 3D virtual environment to \removed{increase immersion and}make the experiment more similar to a real teleoperation experience.
We collected subjective feedback to estimate the perceived workload during the experiment, through the NASA-TLX test \cite{hart_development_1988}, after the first and last run of each trial.
Moreover, participants were asked to fill a subjective feedback questionnaire, as in \cref{t:quest_learn}.
We considered three metrics to evaluate the teleoperation success: time of completion, travelled distance, and number of collisions.

\begin{figure}[h]
\begin{subfigure}{\columnwidth}
  \includegraphics[width=\columnwidth]{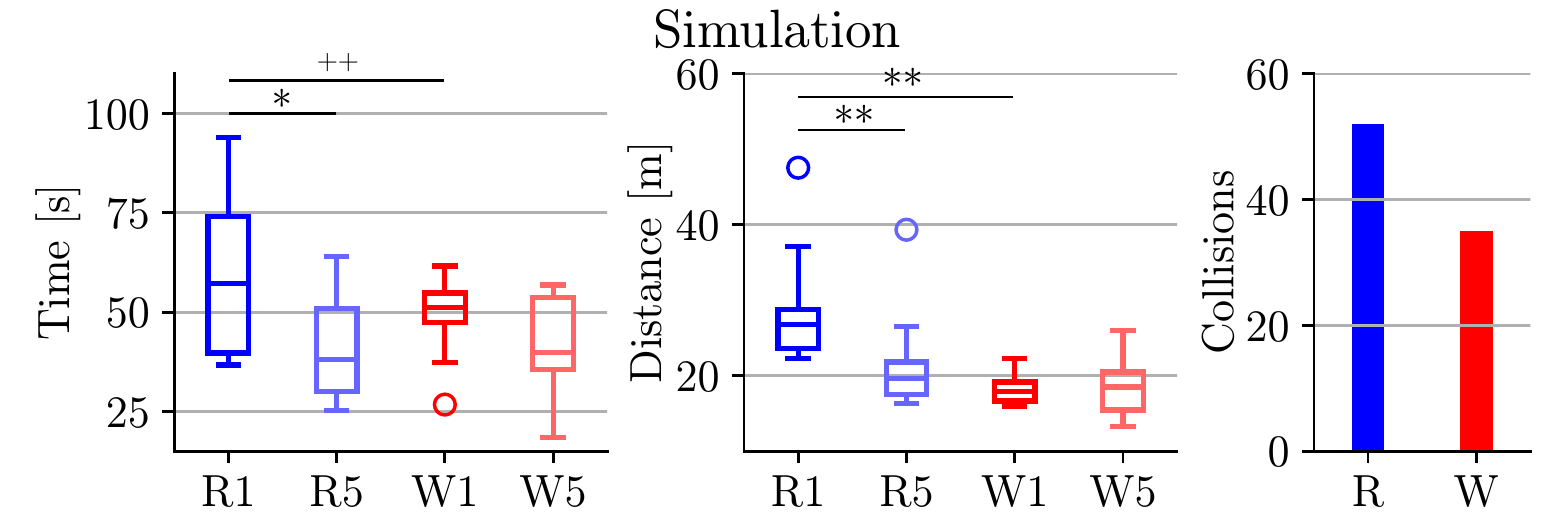}
\end{subfigure}\\
\begin{subfigure}{\columnwidth}
  \includegraphics[width=\columnwidth]{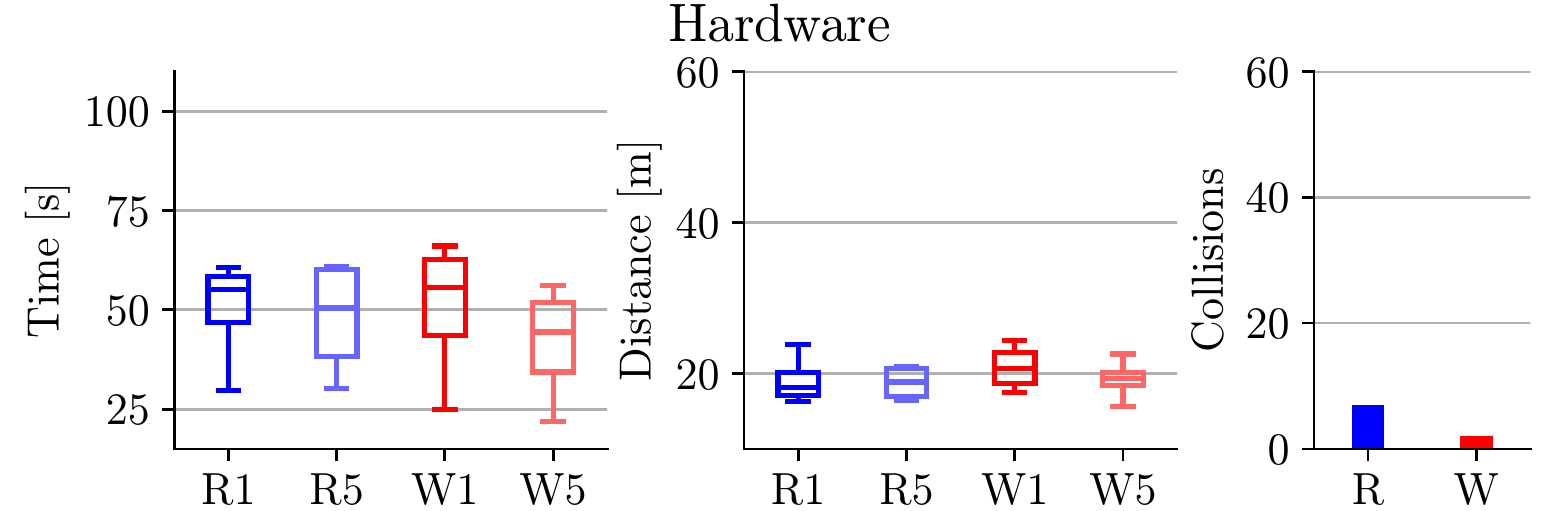}
\end{subfigure}\\
    \caption{Time, distance, and collisions for Experiments II-III, for remote users (R) and wearable interface users (W), runs 1 (R1,W1) and 5 (R5,W5). In simulation, results show a significant improvement in time for Group R, which is not present for Group W, demonstrating the faster performance of Group W from the first run. Group W also showed a significantly reduced traveling distance in the first run and fewer overall collisions. In hardware, results are less evident for the first two metrics, while the collision number is consistent.  ($\text{*}p<0.05, \text{*}\text{*}p<0.01, ^{++}p<0.01$ for F-Test)}
    \label{f:res_learn}
\end{figure}

\textbf{III - Learning the motion-based interface (hardware): }
\removed{As a more practical validation step, we implemented a real-world version of our simulation in order to assess the participants' ability to acquire the same skills as in the simulation task.}
4 subjects controlled the flight of a quadrotor through six gates in the MoCap room, in a setup similar to~\cref{f:path}\mod{, in order to qualitatively validate the results obtained in simulation}.
The quadrotor commands were set to reproduce the hand motion with a scaling factor of 6 to prevent instabilities.
We run our experiments on a Bitcraze Crazyflie 2.1 quadrotor. 
During the experiments, we approximated the quadrotor dynamics with the ones of a point-mass (as often done in the literature \cite{schiano2018dynamic}).
\rem{A ground control station running ROS was established with a dedicated radio channel to issue commands to the drone and receive its sensor data.}\\
\textit{\textbf{Wearable interface shows promising results compared to a remote controller: }}
\cref{f:res_learn} shows experimental results.
In simulation, during the first run, \removed{participants using the wearable interface} \mod{Group W} presented a smaller time variance ($p < 0.01$) than \removed{remote users} \mod{Group R}.
During the last run, performance is comparable between the two interfaces, but while Group W shows no significant improvements, Group R performed significantly better ($p = 0.047$) from the first run. 
We can see a substantial difference in remote users with training, while wearable users can perform in less time from the start.
Also, the traveled distance shows substantial differences: in the first attempt, Group W traveled over shorter paths ($p < 0.01$) than Group R, saving on average $33\%$ of travelling distance. Similarly, remote users had to get used to the interface to improve their performance from the first to the last attempt ($p < 0.01$).
Finally, Group W incurred in fewer environment collisions (52 vs 35). This can be due to the more natural control interface, similar to a common manipulation task.
Time and distance metrics do not show a significant difference in the hardware scenario\rem{, which can be due to the limited maximum speed imposed on the robot in order to prevent damage}. Nonetheless, wearable interface users collided with the obstacles fewer times, showing improved control over the drone trajectory.

\begin{figure}[h!]
\begin{subfigure}{\columnwidth}
  \includegraphics[width=\columnwidth]{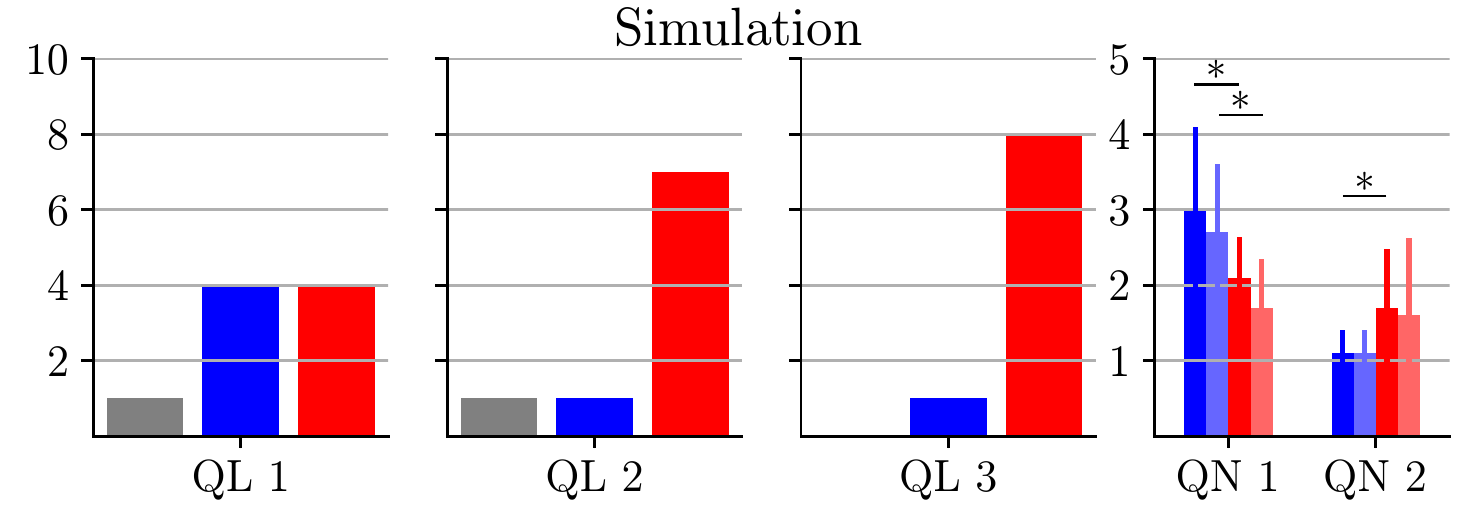}
\end{subfigure}\\
\begin{subfigure}{\columnwidth}
  \includegraphics[width=\columnwidth]{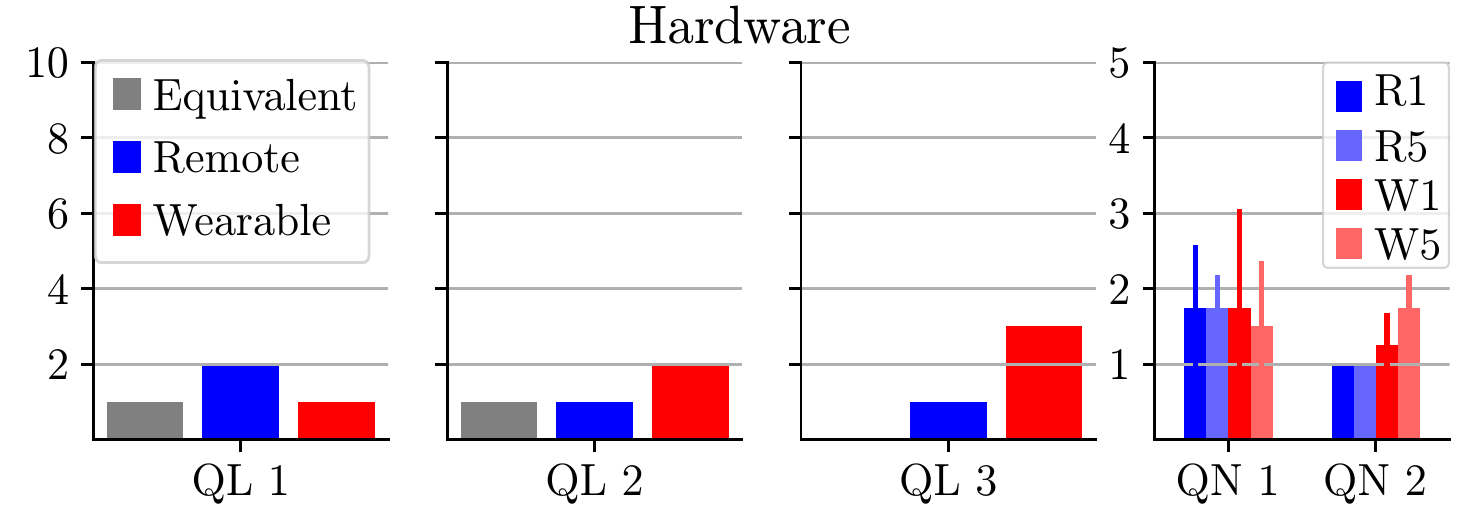}
\end{subfigure}\\
\caption{Subjective survey (QL 1-2-3) and NASA-TLX (QN) test response for Experiments II-III. Most participants found the proposed solution easier to use after training and preferred it over the remote in both cases. 
Moreover, they found the use of a remote controller more mentally demanding during the whole test and the wearable interface more tiring only in the first run in simulation. ($\text{*}p<0.05$)}
\label{f:learn_quest_HW}
\end{figure}

In \cref{f:learn_quest_HW}, we show the responses to the feedback questionnaire.
After the first run, the population's opinion on the ease of use was equally
spread between remote and motion-based interface\rem{, with one undecided participant}. 
At the end of the experiment, instead, $78\%$ of the participants (7 out of 9) declared to find the motion-based interface easier.
Moreover, 8 of them declared to prefer the proposed interface over a standard remote.
The NASA-TLX test shows significant differences results for questions 1 and 2, respectively, relative to mental and physical workload. 
Participants found significantly more mentally demanding the use of a remote both for the first ($p = 0.033$) and the last ($p = 0.019$) run, and the wearable system more tiring ($p = 0.040$) only in the first run.
In hardware, the feedback responses are coherent, even if obtained from a smaller subject pool\rem{, with the simulation results}.
To the individual feedback question QL 4, six participants declared to prefer the wearable system due to its intuitiveness, in terms of faster operation confidence and less time spent thinking about ``how to use it''. Also, three subjects defined it as a more engaging solution.

\begin{table}[th]
\begin{center}
\begin{tabular}{ | l | l |}
 \hline
 \textbf{QL 1} & Which interface was easier to use in the FIRST run? \\
 \hline
 \textbf{QL 2} & Which interface was easier to use in the LAST run? \\
 \hline
 \textbf{QL 3} & Which interface did you prefer? \\
 \hline
 \textbf{QL 4} & Why? \\
 \hline
\end{tabular}
\caption{Subjective feedback form for Experiments II-III}
\label{t:quest_learn}
\end{center}
\end{table}

\begin{figure}[h]
\begin{subfigure}{.33\columnwidth}
  \centering
  \includegraphics[width=.95\linewidth]{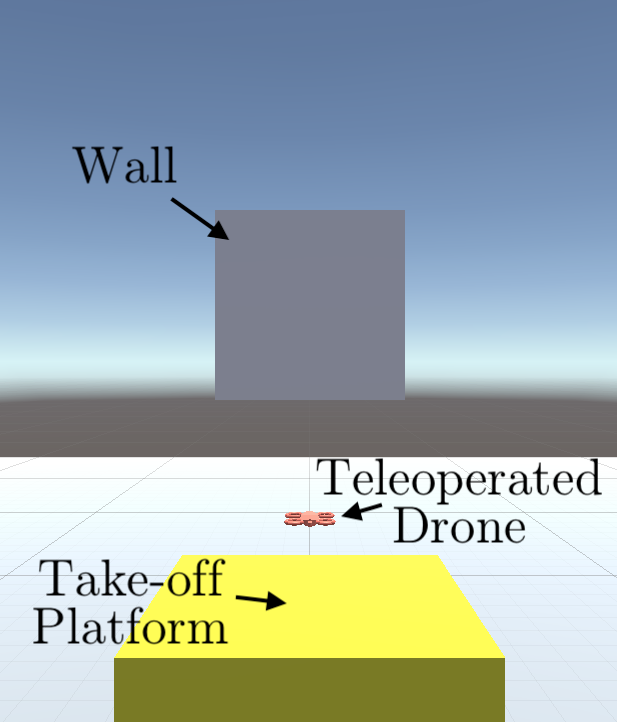}
\end{subfigure}%
\begin{subfigure}{.33\columnwidth}
  \centering
  \includegraphics[width=.95\linewidth]{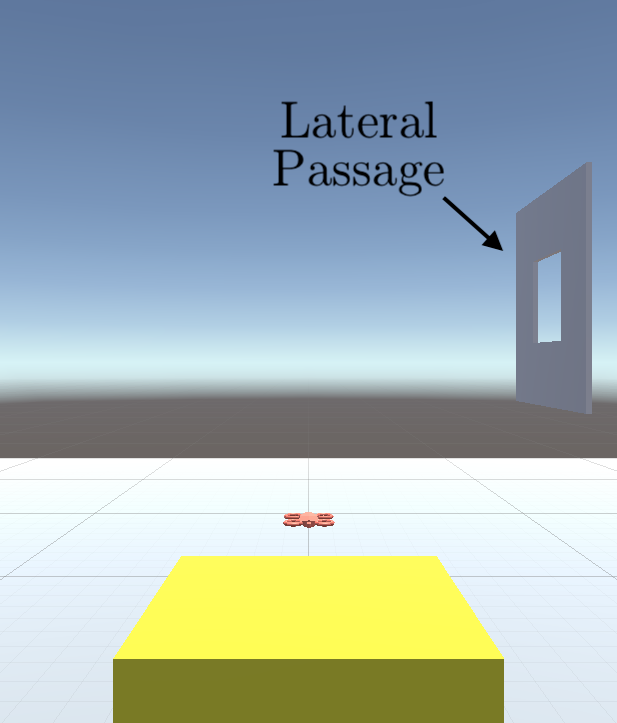}
\end{subfigure}
\begin{subfigure}{.33\columnwidth}
  \centering
  \includegraphics[width=.95\linewidth]{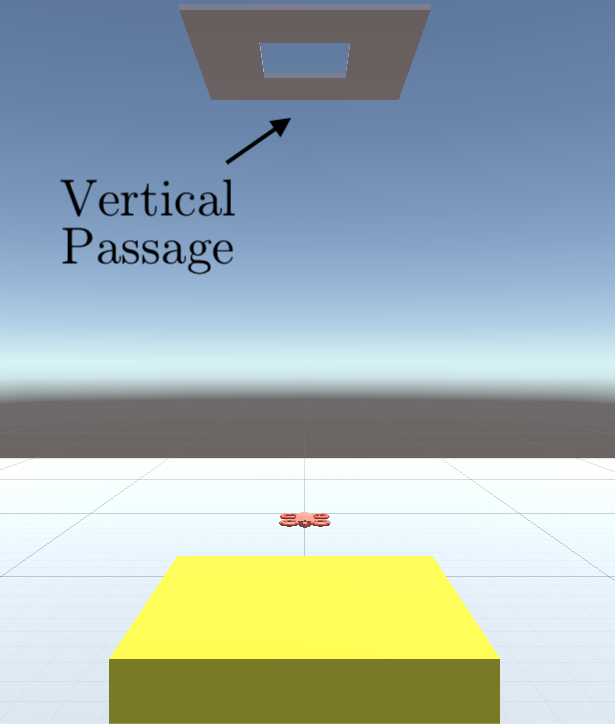}
\end{subfigure}
\caption{Simulation environment for the haptics task. In the figure the three different experimental conditions are shown (Task 1: approach a wall in front of the user, Task 2: go through a lateral passage and Task 3: go through a vertical passage).}
\label{f:scene_hapt}
\end{figure}

\textbf{IV - Haptic navigation test (simulation): }
\mod{9 participants were recruited to test the effectiveness of the haptics in augmenting the user's perception of the environment.}
We specifically wanted to address and correct depth perception errors~\cite{walk_comparative_1961}, commonly affecting teleoperation \cite{chen_human_2007}.
\mod{We consider condition R vs W vs H.}
Instead of operating the drone through a path, the participant was asked to perform three specific tasks depicted in \cref{f:scene_hapt}.
Task 1 consists of approaching a flat object, placed in front of the user, as much as possible without colliding with it.
We chose this scenario because it is particularly hard for an observer to accurately estimate the distance from a featureless object in front of them.
Nonetheless, ordinary tasks such as inspection and navigation in confined environments require to approach walls to examine them while preventing collisions~\cite{jimenez2019contact}.
Task 2 and 3 consist, respectively, of going through a lateral or vertical opening, placed perpendicularly to the user's sight.
Shadows rendering was deactivated in the simulator for this stage, as they can improve a user's perception skills \mod{\cite{puerta_power_1989}}.
This is a plausible assumption: during teleoperation, illumination conditions and altitude can affect or cancel shadows from the sight.
The choice of this task is also related to the common case in which an operator is required to steer a drone through an opening, which is in LoS (e.g., passage through a window or a door).
The first device to be used was chosen in a pseudo-random fashion.
Each subject was allowed to train for one minute for each task.
After training, they performed each task five times consecutively.
At the end of each task, each participant filled the NASA-TLX test.
Moreover, participants were asked to fill a subjective feedback questionnaire consisting of 4 questions, as in \cref{t:quest_hapt}.

\newpage

\begin{figure}[t]
\begin{subfigure}{\columnwidth}
  \includegraphics[width=\columnwidth]{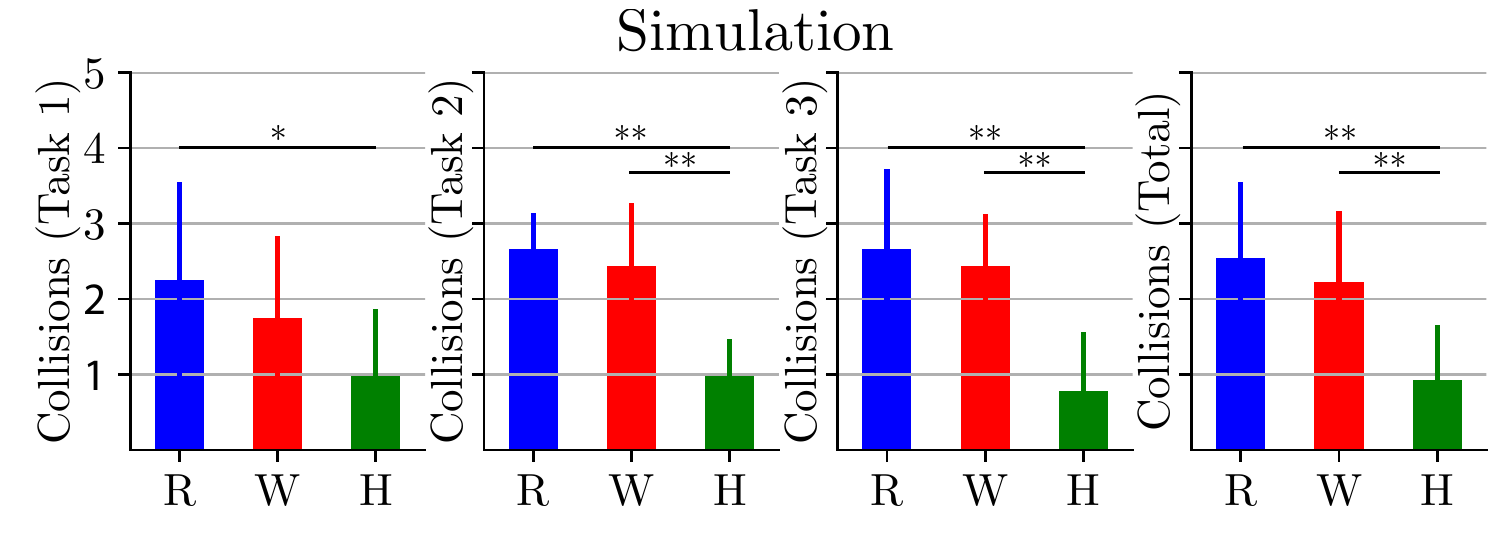}
\end{subfigure}\\
\begin{subfigure}{\columnwidth}
  \includegraphics[width=\columnwidth]{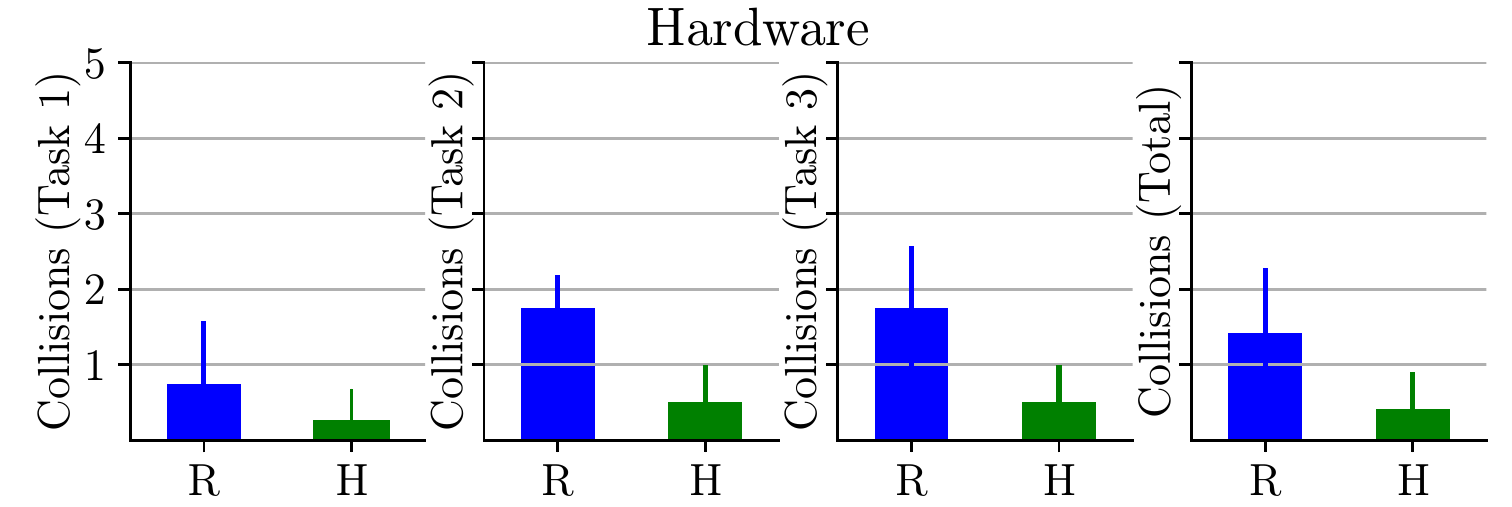}
\end{subfigure}\\
\caption{Collisions per subject during the haptics test (Experiment IV-V), \removed{using a remote controller or the proposed wearable haptic interface} \mod{for Groups R,W and H}. Results show a clear advantage in the adoption of vibrotactile feedback \mod{(Group H)} for obstacle avoidance, reducing the number of collisions \removed{for Task 1 and 2 }by more than $50\%$ overall\removed{, and providing non significant improvement for Task 3}. 
These results are consistent in the hardware experiment\removed{, where also an improvement for Task 3 can be appreciated}.
($\text{*}p<0.05, \text{*}\text{*}p<0.01$)}
\label{f:collisions_HW}
\end{figure}

\begin{figure}[t]
\begin{subfigure}{\columnwidth}
  \includegraphics[width=\columnwidth]{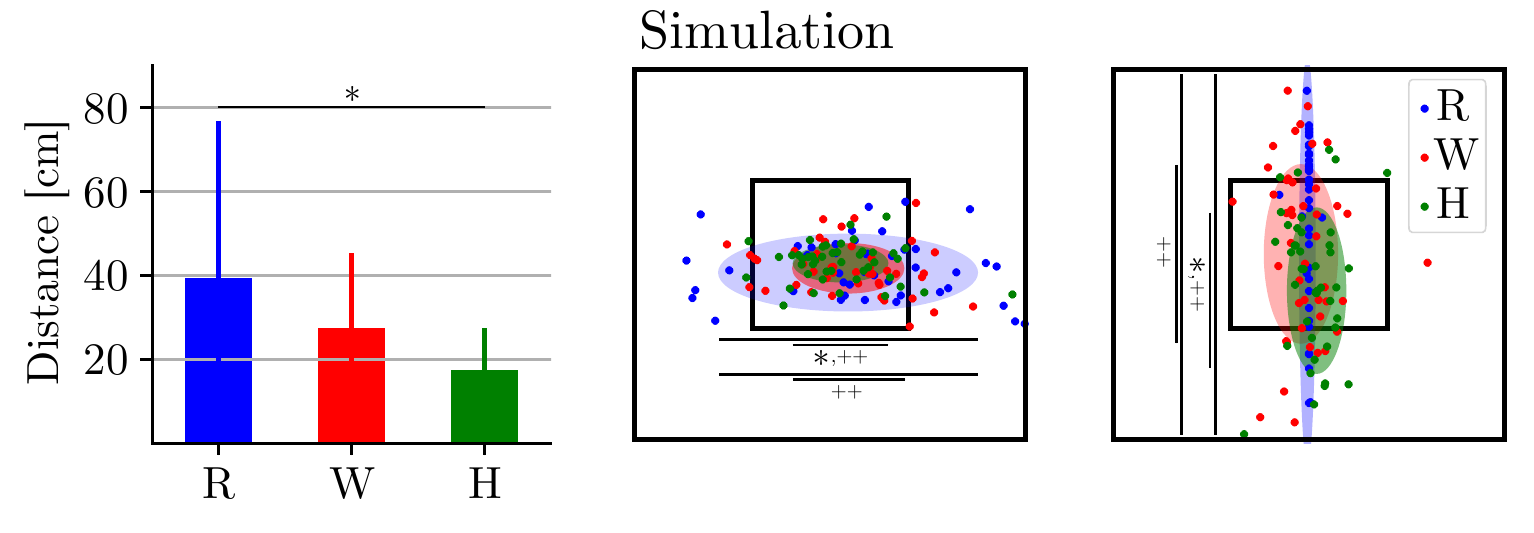}
\end{subfigure}\\
\begin{subfigure}{\columnwidth}
  \includegraphics[width=\columnwidth]{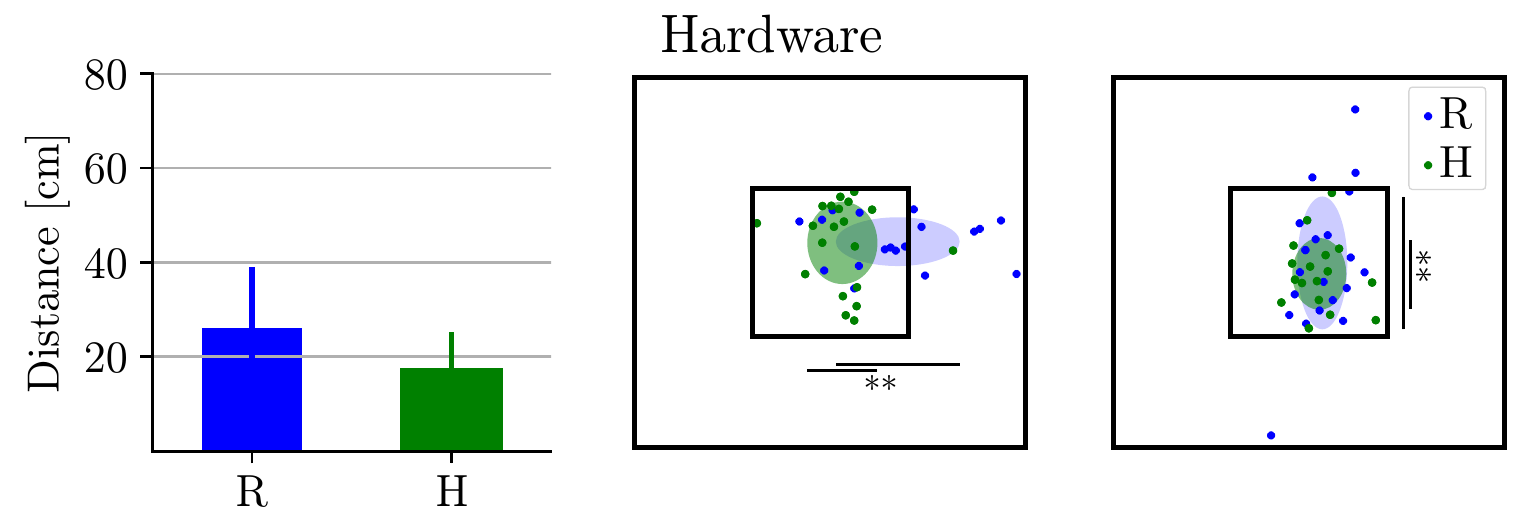}
\end{subfigure}\\\caption{Distances from the front obstacle for Task 1 and crossing position for Tasks 2 and 3. 
\removed{Wearable interface users}\mod{Group H} was able to approach the wall more closely on average.
The distribution of the crossing location shows a significantly lower variance on the axis of visual restriction.
Results are consistent in both simulation and hardware tests.\\
($\text{*}p<0.05, ^{++}p<0.01$ for F-Test)}
\label{f:cross}
\end{figure}

\begin{figure}[th]
\begin{subfigure}{\columnwidth}
  \includegraphics[width=\columnwidth]{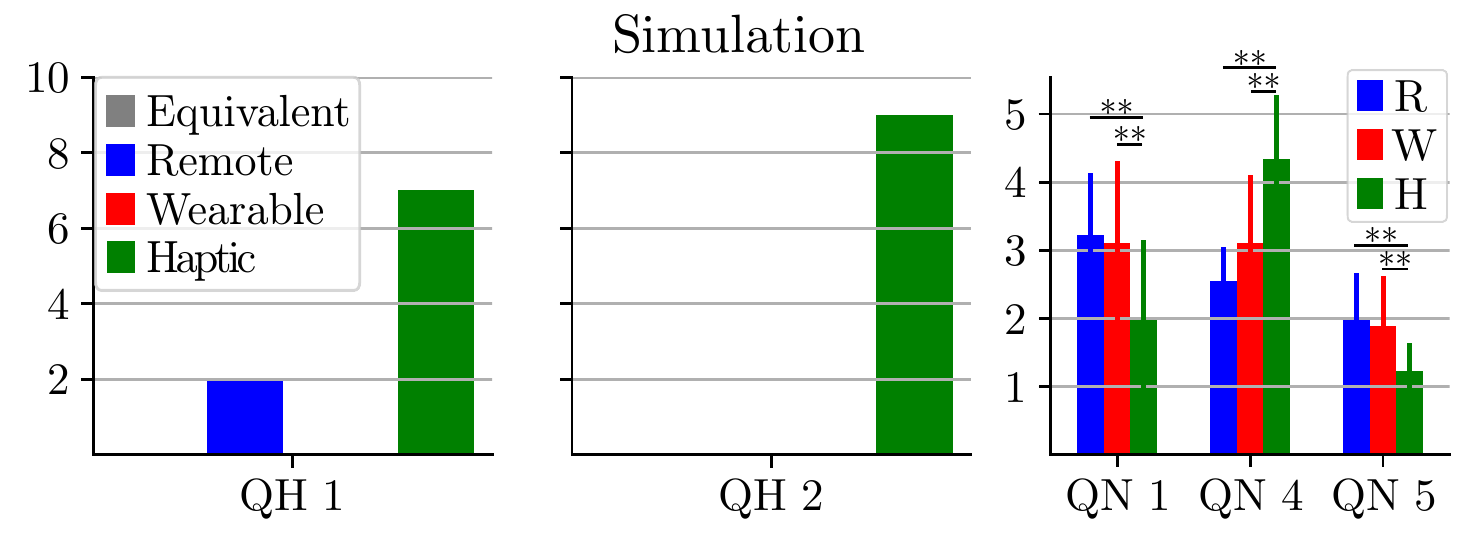}
\end{subfigure}\\
\begin{subfigure}{\columnwidth}
  \includegraphics[width=\columnwidth]{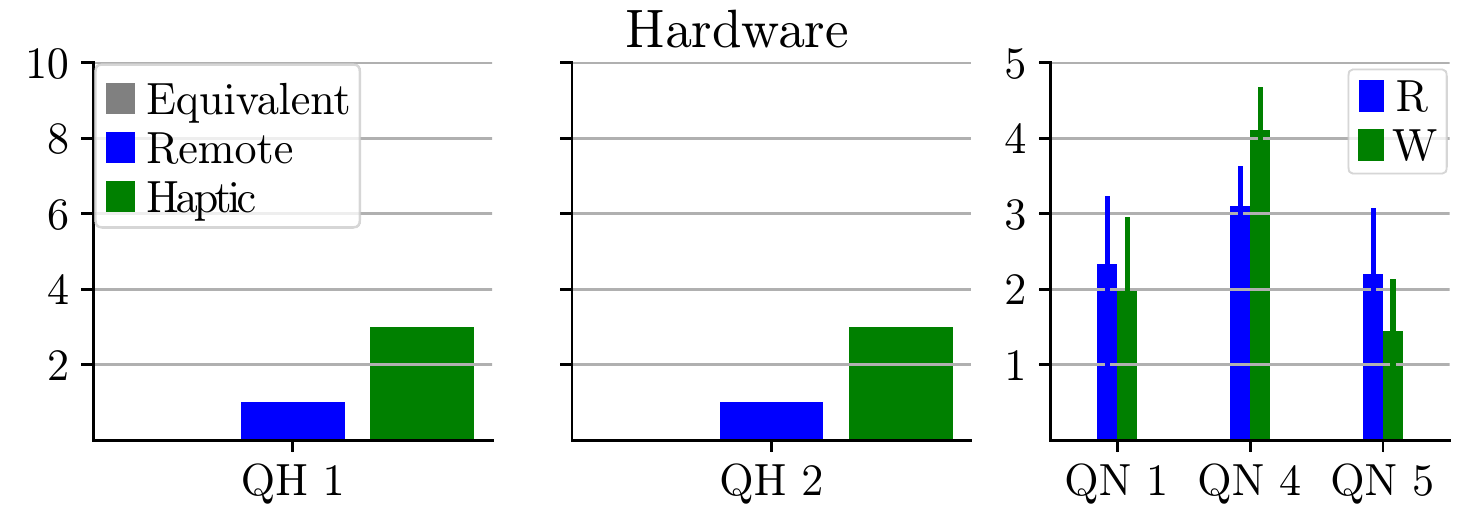}
\end{subfigure}\\
\caption{Subjective survey (QH 1-2) and NASA-TLX (QN) responses for Experiments IV-V. Participants consistently preferred the \removed{waerable}\mod{haptic} interface, and found it less mentally demanding, perceived to have achieved better results and felt less stressed and insecure with respect to an interface not providing tactile cues.
($\text{*}p<0.05, \text{*}\text{*}p<0.01$) }
\label{f:quest_hapt_HW}
\end{figure}

\textbf{V - Haptic navigation test (hardware):}
Similarly to \mod{experiment III,} \mod{4  new subjects  were asked to control the drone in a real-world implementation of the three scenarios depicted in~\cref{f:scene_hapt}. Here, only conditions R and H were tested.}
\removed{implemented the haptic navigation tasks in the MoCap room for hardware tests, using the same instrumentation as in experiment II. 
Here,  4  new subjects  controlled  the  flight  of  a quadrotor}\\
\textbf{\textit{Haptic feedback improves performance in the given tasks: }} 
\cref{f:collisions_HW} illustrates the number of collisions that occurred with the environment during the three phases of the experiment. 
\removed{In simulated Tasks 1 and 2} \mod{In each task}, participants using the proposed interface had less than half the accidents of remote users ($p = 0.036$ in Task 1, $p < 0.01$ in Tasks 2,3).
\removed{During Task 3, the participants using the haptic glove had less collisions but not at a significant level.}
In the real-world test, results are coherent with what seen in simulation\removed{, with a visible improvement also in Task 3}.
In \cref{f:cross}, we show the approach distance to the front obstacle and crossing point during the tests\removed{, which are similar to the results in the simulated environment}.
\cref{f:cross} (left) summarizes the average wall distance for the \removed{two} \mod{three} cases. 
Here, \mod{Group H} outperformed \removed{remote users} \mod{Groups R and W} and could reach a closer position to the wall while preventing collisions ($p_{RH} = 0.035$).
\cref{f:cross} (center, right) illustrates the distribution of the drone position while crossing the lateral and vertical gates, respectively.
Independently from the success rate, the use of \removed{haptic} \mod{the glove (Groups W and H)} provided a smaller variance both in simulation ($p_{RW}, p_{RH} < 0.01$) and hardware on the X-axis, corresponding to the frontal facing orientation of the participant.
In \cref{f:quest_hapt_HW}, we show the responses to the final feedback questionnaire.
\removed{All} \mod{Most of the} participants agreed on the superiority of the haptic interface for the three tasks, and all participants preferred this solution over the remote.
The effectiveness of the haptic feedback was given a score of $8.7/10$ in simulation and $9.7/10$ in hardware.
Here, the NASA-TLX test showed significant differences results for questions 1, 4, and 5, respectively relative to mental workload, perceived success in the task, and feeling of insecurity and stress.
Participants found the test less mentally demanding, perceived to have achieved better results on average, and felt less stressed and insecure ($p_{RW}, p_{RH} < 0.01$).
During the hardware test, the responses to our feedback survey were coherent with the ones obtained in simulation.
The final individual feedback question QH 4 showed results similar to QL 4. 
Four subjects mentioned the intuitiveness and engaging appeal of the haptic device. 
Nearly everyone (11/13 subjects) felt that the proposed device strongly aided their depth perception.

\begin{table}[th]
\begin{center}
\begin{tabular}{ | l | l |}
 \hline
 \textbf{QH 1} & Which interface was easier to use for the task? \\
 \hline
 \textbf{QH 2} & Which interface did you prefer? \\
 \hline
 \textbf{QH 3} & How useful was the haptic inteface? \\
 \hline
 \textbf{QH 4} & Why? \\
 \hline
\end{tabular}
\caption{Subjective feedback form for Experiments IV-V}
\label{t:quest_hapt}
\end{center}
\end{table}

\section{Conclusions}
This work presents a systematic approach used to investigate the viability of controlling a drone with a haptic-enabled data glove with respect to a canonical user interface, such as remote controllers.
The motion-based interface allows a natural control over the robot's trajectory, and the tactile feedback conveys information about the environment surrounding the drone. 
This choice is finalized to provide an augmented sensory stream to the operator and allow them to efficiently avoid obstacles during flight, preventing the need for an automatic collision avoidance system.
The haptic glove was designed, implemented, and evaluated in three subsequent steps. First, the haptic feedback was tested. It was shown that users can accurately ($98\%$ of the times) identify the cue direction on their hand thanks to the placement of six tactors in strategical positions of the glove. Subsequently, we investigated the learning ability of users approaching for the first time this system. We realized that, in a simulated environment, this interface can outperform a standard remote, providing a shorter time for completing a mission, a lower traveled distance, and fewer occurred collisions. This, despite the fact that most users had prior experience in the use of remote controller, and no training with the proposed one. Finally, we let participants test the haptic feedback function in three different simulated tasks, where the drone was set to fly close to obstacles and in confined spaces. The experiment demonstrated that adding haptic information to simple visual feedback in LoS significantly improves the success of such missions\mod{, making the user able to fly in closer proximity with obstacles without colliding with them}. 

Once the design of the glove was validated and tested in simulation, it could be tested in a real environment.
We reproduced our experiments in a MoCap room with an additional set of participants and observed that the wearable interface can be a viable substitution to a remote for the teleoperation of a real drone. 
In this case, the augmented feedback from the haptic device showed convincing improvements also when working on the real quadrotor.
Participants stated in a subjective feedback survey that they found the proposed interface more intuitive and easy to learn, more engaging, and that it improved their perception of depth through haptics.

This work is a first step towards the design and implementation of motion-based, haptic-enabled HRIs, and it opens to important questions towards exciting future directions. 
\mod{First, validating our system on a larger pool of subjects would surely be beneficial to provide greater insight of its effectiveness.}
We are currently working to integrate our system and making it smaller to be easily worn by users.
Also, our clutch mechanism was perceived as nonintuitive by some participants in the initial stage. 
One way to free the users from using a clutch system would be to switch from mapping the hand pose from position to velocity drone commands.
The human-robot mapping, as it is implemented at the moment, is a simple scaling of the hand position.
It could be improved to data-driven mappings, as in \cite{khurshid_data-driven_2015} to help the users to learn it faster. 
Moreover, to allow the deployment of our system in environments in which a MoCap is not available, the use of wearable sensors to track hand motion should be investigated.
\pagebreak

\bibliographystyle{IEEEtran}
\bibliography{bib/alias,bib/IEEEConfAbrv,bib/IEEEabrv,bib/otherAbrv,bib/bibCustom}

\end{document}